\documentclass[conference]{IEEEtran}

\usepackage{wrapfig}

\usepackage{longtable}

\usepackage{booktabs}

\usepackage{color, colortbl}
\definecolor{Gray}{gray}{0.9}
\usepackage{subfig}
\usepackage{comment}
\usepackage{array}
\newcolumntype{L}[1]{>{\raggedright\let\newline\\\arraybackslash\hspace{0pt}}m{#1}}
\newcolumntype{C}[1]{>{\centering\let\newline\\\arraybackslash\hspace{0pt}}m{#1}}
\newcolumntype{R}[1]{>{\raggedleft\let\newline\\\arraybackslash\hspace{0pt}}m{#1}}

\usepackage{comment}
\usepackage{multirow} 

\usepackage{amsmath} 
\usepackage{amssymb}  
\usepackage{lipsum}
\usepackage{adjustbox}
\newcommand{\xhdr}[1]{{{\bf #1.}}}

\graphicspath{{figures/}}

\usepackage{cite}

\usepackage{amsmath}
\usepackage{algorithmic}

\usepackage{array}

\usepackage{url}

\usepackage{flushend}

\IEEEoverridecommandlockouts

\begin{document}

\title{Deep Learning Assessment of Tumor Proliferation in Breast Cancer Histological Images}

\author{

\IEEEauthorblockN{Manan Shah}
\IEEEauthorblockA{The Harker School\\
San Jose, CA, USA\\
manan.shah.777@gmail.com}

\and

\IEEEauthorblockN{Christopher Rubadue, David Suster, Dayong Wang}
\IEEEauthorblockA{Beth Israel Deaconess Medical Center\\
330 Brookline Avenue, Boston, MA, USA \\
\{crubadue, dsuster, dwang5\}@bidmc.harvard.edu}
}

\maketitle

\begin{abstract}
Current analysis of tumor proliferation, the most salient prognostic biomarker for invasive breast cancer, is limited to subjective mitosis counting by pathologists in localized regions of tissue images. This study presents the first data-driven integrative approach to characterize the severity of tumor growth and spread on a categorical and molecular level, utilizing multiple biologically salient deep learning classifiers to develop a comprehensive prognostic model. Our approach achieves pathologist-level performance on three-class categorical tumor severity prediction. It additionally pioneers prediction of molecular expression data from a tissue image, obtaining a Spearman's rank correlation coefficient of 0.60 with \textit{ex vivo} mean calculated RNA expression. Furthermore, our framework is applied to identify over two hundred unprecedented biomarkers critical to the accurate assessment of tumor proliferation, validating our proposed integrative pipeline as the first to holistically and objectively analyze histopathological images.
\end{abstract}

\IEEEpeerreviewmaketitle

\section{Introduction}
Breast cancer is the most common cancer in women worldwide, with over 1.2 million new cases diagnosed in 2012 \cite{ma2013breast}. Cancer assessment is influenced by environmental and clinical factors, but it is universally accepted that tumor proliferation speed (tumor growth) is an important biomarker representative of progression rate and outcomes \cite{veta2016mitosis, cheang2006immunohistochemical}. Specifically, high proliferation speed is associated with worse outcomes \cite{veta2016mitosis}. The assessment of this biomarker critically influences patient treatment plans, allowing for patients with more aggressive tumors to be treated with the corresponding therapy \cite{cheang2006immunohistochemical}. 

In a clinical setting, tumor proliferation is manually assessed by pathologists under a regime of counting mitotic figures in hematoxylin \& eosin (H\&E) stained histological slide preparations that are examined under a high powered microscope. Although ubiquitous, the process of counting mitoses has been reported to suffer from reproducibility problems that reflect the underlying subjectivity of the process \cite{veta2016mitosis}. In addition, the simple methodology of pathologist mitosis counting and subsequent thresholding fails to account for pathological features including tumor extent, tissue density, and relative mitosis density. The mean expression of eleven prognostic RNA sequences, an objective measure of the proliferation score, requires extensive \textit{ex vivo} molecular tests, relegating current practices to an inadequate and subjective pathologist diagnosis \cite{cheang2006immunohistochemical, nielsen2010comparison}.

This study introduces a comprehensive deep learning-based pipeline constructing and unifying models across several associated tasks of tumor localization, mitotic figure identification, and high-level feature extraction to classify categorical tumor grades (0--2) and predict RNA proliferation speed scores from histological whole slide images (WSIs). Furthermore, we aim to identify salient biomarkers related to tumor diagnosis to serve as the basis for future studies. The data-driven integrative approach presented here is generalizable and will be useful to analyze other cancerous tumors.

\section{Methods}

\begin{figure*}[tbh]
\centering
\includegraphics[width=\textwidth]{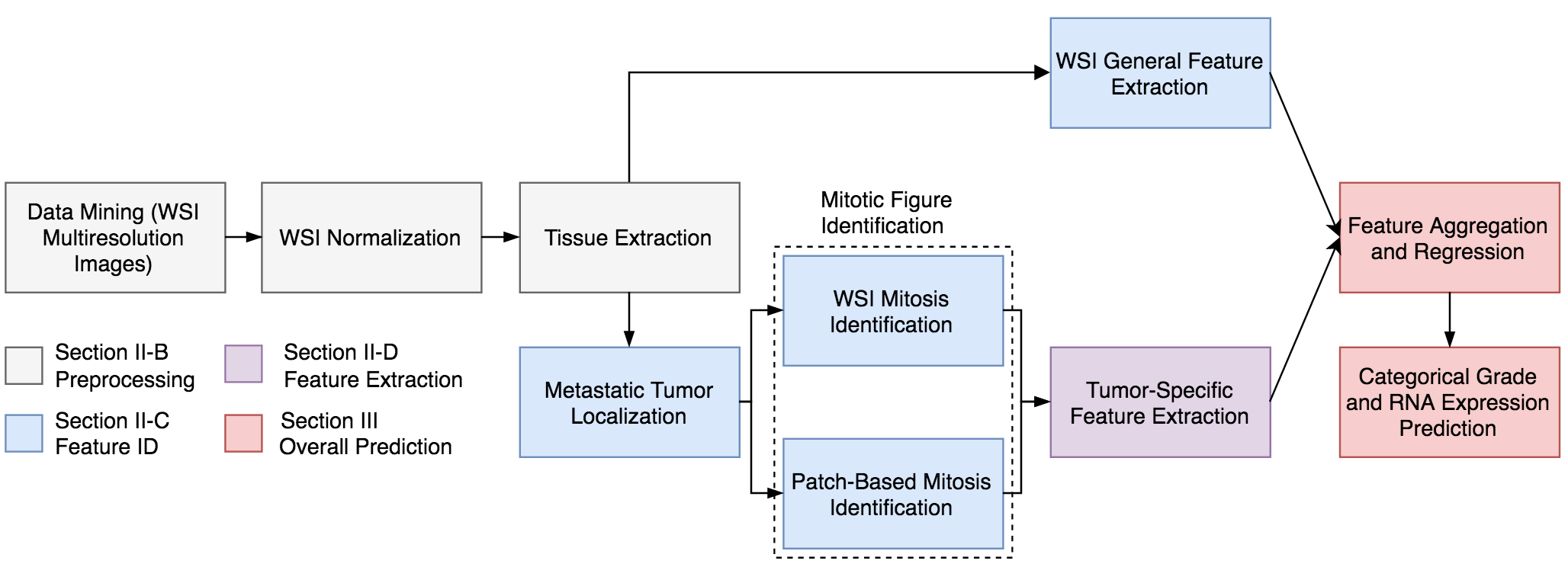}
\caption{\textit{Analysis Process for Tumor Proliferation}. Each multiresolution input WSI is sequentially processed with a series of biologically salient steps including tissue extraction, tumor localization, and mitotic figure tumor-specific feature extraction. These processes are combined with general extracted features to generate molecular and categorical predictions. }
\label{pipeline}
\end{figure*}

\subsection{Dataset Description}

Three datasets were used in this study to train primary and auxiliary models \cite{IMAGe}. Our primary evaluation dataset consisted of 500 whole slide images with magnification levels from 10$\times$ to 40$\times$ that are annotated with a tumor score based on mitosis counting by pathologists and a molecular (RNA) proliferation score \cite{IMAGe}. We additionally use 73 2 $\text{mm}^2$ magnified images annotated with mitotic figures and 148 WSIs partially annotated with tumor (high cellularity) regions for supplemental training. 

\subsection{Data Mining, WSI Normalization, and Tissue Extraction} 

As our WSIs originate from three international pathology centers, each exhibited different staining methods. To ensure that such variations in the color and intensity of H\&E staining would not hamper the effectiveness of subsequent quantitative image analysis, we employed the Bejnordi et al's WSI Color Standardization (WSICS) procedure \cite{bejnordi2016stain}. Normalizing each WSI and images from auxiliary datasets ensured that our subsequent methods exhibited stain invariance between tissue preparation methods. Tissue regions were next extracted from each stain standardized input WSI using Otsu's method on a HSV representation of the original RGB image \cite{sural2002segmentation, otsu1975threshold}. We subsequently removed small artifacts and expanded remaining regions via binary dilation to obtain a holistic tissue mask.

\subsection{Network Construction} \label{netconstruct}
Our pipeline (Fig. \ref{pipeline}) next performed three tasks: metastatic tumor localization, mitotic figure identification, and WSI general feature extraction. We used a magnification level of 10$\times$ for the tumor localization process to identify high-level patterns, and we conducted mitosis identification and feature extraction on the 40$\times$ level for detailed analysis. 

\subsubsection{Metastatic Tumor Localization} \label{tumorlocalization}

Having extracted and normalized tissue regions from each input WSI, it is important to identify candidate regions for mitotic activity indicative of tumor proliferation. Such regions are biologically characterized as high-cellularity areas with proliferative activity often represented at the edges of the tissue abnormality. 

Fig. \ref{models} depicts the four employed network architectures, the first three of which are recognized as state-of-the-art convolutional neural networks (CNNs) for object recognition \cite{szegedy2015going, he2015deep, simonyan2014very}. Each CNN generated output tumor probability heatmaps with a ``sliding-window'' approach \cite{szegedy2013deep}, classifying overlapping tissue patches for tumor probability and assigning the resulting value to the center pixel of each patch. The fourth network, which we named LocNet, reduced the number of free parameters and operated on a fully convolutional paradigm \cite{long2015fully}. LocNet allowed for arbitrarily sized inputs (we use $1\text{k} \times 1\text{k}$ patches) and produced downsampled corresponding heatmaps for each patch as opposed to singular probability outputs. We resized and stitched these probability heatmaps over each WSI to rapidly generate a comprehensive heatmap. 

Network training was framed as an active learning problem \cite{cohn1996active} as each annotated image contained a non-exhaustive list of tumor regions. We, therefore, separated the process into two components, with the first stage defining annotated patches as positive and identifying a random sample of remaining tissue patches as negative. Heatmaps were subsequently produced using each model and additional regions predicted as positive with over 95\% confidence were appended to the initial positive training set. All models were retrained with the refined data; subsequent second stage results better eliminated regions misclassified in the first stage. The trained models were used both to identify tumor regions within which to perform mitotic figure identification and to provide informative features regarding tumor shape, density, area, extent, and location. 

\begin{figure*}[bt]
\small
\centering
\begin{tabular}{ L{3.4cm} L{2cm} L{1.7cm} L{1cm} L {2cm} L{2cm} L{2.5cm}}
\toprule
\textbf{Network} & \textbf{\# Layers} & \textbf{\# Params} & \textbf{RF} & \textbf{Input} & \textbf{Output} & \textbf{Propagation Time} \\
\midrule \midrule
\multicolumn{7}{c}{Metastatic Tumor Localization (\ref{tumorlocalization})}\\ \midrule
GoogLeNet \cite{szegedy2015going} & 27 & 5.97 M & 49 & $224 \times 224$ & 1 $\times$ 2 & 562.62 ms \\
ResNet-34 \cite{he2015deep} & 34 & 13.9 M & 49 &  $256 \times 256$ & 1 $\times$ 2 & 204.55 ms \\
VGG-13 \cite{simonyan2014very} & 39 & 134.3 M & 9 & $224 \times 224$ & 1 $\times$ 2 & 208.85 ms  \\
LocNet$^*$ & 12 & 4.55 M & 9 & 1k $\times$ 1k & $63 \times 63$  & 77.47 ms\\  \midrule
\multicolumn{7}{c}{Mitotic Figure Identification (\ref{mitoticidentification})}\\ \midrule
DenseNet \cite{huang2016densely} & 118 & 1.02 B & 9 & $224 \times 224$ & 1 $\times$ 2 & 500.09 ms \\
GoogLeNet \cite{szegedy2015going} & 27 & 5.97 M & 49 & $224 \times 224$ & 1 $\times$ 2 & 562.62 ms \\
GoogLeNet FCN$^*$ & 27 & 5.97 M & 49 & 1k $\times$ 1k & $48 \times 48$ & 340.53 ms \\
LocNet$^*$ & 12 & 4.55 M & 9 & 1k $\times$ 1k & $63 \times 63$  & 77.47 ms\\ 
MitosNet$^*$ & 6 & 21.1 K & 16 & 1k $\times$ 1k & $63 \times 63$ & 44.21 ms\\ \midrule
\multicolumn{7}{c}{General Feature Extraction (\ref{generalextraction})}\\ \midrule
Tumor + 3C/P$^*$ & 18 & 5.16 M & 64 & 1k $\times$ 1k & $1 \times 3$ & 492.4 ms  \\
Mitosis + 3C/P$^*$ & 12 & 5.93 M & 64 & 1k $\times$ 1k & $1 \times 3$ & 143.1 ms  \\
\bottomrule
\end{tabular}
\caption{\textit{Network Architectures}. Parameters and statistics characterizing the seminal networks developed in Section 2. The propagation time column characterizes the mean forward pass time for one input image, and the RF column represents the receptive field (in pixels) of each network. $^*$\textit{New networks specifically developed for this work; C/P indicates Convolution/Pooling layers.}}
\label{models}
\end{figure*}

\subsubsection{Mitotic Figure Identification} \label{mitoticidentification}

We next constructed mitotic figure detectors to identify biologically salient features within tumor areas. Due to aberrant tumor chromosomal makeup, mitotic figure appearances may vary from typical examples of hyperchromatic objects with an absence of a clear nuclear membrane and hairy protrusions around edges. 

Current state-of-the-art methods in the field of computational mitosis identification are trained and evaluated on high quality, standardized, and localized tumor regions with well-defined mitotic figures \cite{veta2015assessment, tek2013mitosis, cirecsan2013mitosis}. However, such methods fail to generalize to our WSI dataset as they often simply learn standardized color and texture filters from their homogeneous training dataset. Prior methods are additionally unable to rapidly generate mitosis identification results over an entire whole slide image. Here, we introduce and apply robust color, texture, and scale invariant mitosis detection networks that rapidly identify mitoses on individual patches and WSIs.

The five networks employed are depicted in the middle section of Fig. \ref{models}. DenseNet, requiring the most parameters, constructed repeated connections between network layers to develop a robust approach. Although the GoogLeNet architecture and its modified fully convolutional counterpart performed reasonably well, the additional complexity encoded within the network architecture excessively distilled the already small mitotic figures. To remedy this issue, we applied the LocNet model and developed a specialized architecture called MitosNet. LocNet and MitosNet exhibited fewer (yet more fine-tuned) layers, capturing the variance between mitoses without degrading effective inference. 

To reduce the false positive rate for identified mitoses, we followed a two-stage training procedure similar to Section \ref{tumorlocalization} using a dataset of pathologist annotated mitotic figures. We initially identified the locations and areas of all nuclei using morphological methods. We defined positive training patches as those nuclei annotated as mitotic, and we identified a random sample of other nuclei as negative. After the first stage of training and output heatmap generation, we subjected our mitosis identifications to further pathologist evaluation and subsequently retrained our models accounting for initially misidentified instances. The resulting robust trained models were used to characterize mitoses in terms of spatial distribution and shape-specific attributes; the process of extracting these features is detailed in Section \ref{tumorextraction}.  

\subsubsection{WSI General Feature Extraction} \label{generalextraction}

In addition to developing methods for the identification of anatomical structures in tumor severity analysis, we created end-to-end networks that predict the output categorical severity grade of the whole slide image from individual patches. These developed networks are defined in the last section of Fig. \ref{models}; each model utilizes outputs of tumor and mitosis networks to extract detailed computational features. Patches are extracted from original WSIs and input to the first network (with static weights) which computes coarse features corresponding to either mitosis identification or tumor localization. Convolutions from the first network's feature volume are next mapped to the input of a second network (with dynamic weights). The second network performed categorical predictions and extracted WSI features. These 1,024 features, combined with 3 predicted class probabilities, were incorporated in the final predictive model. 
\vspace{-3pt}
\subsection{Tumor-Specific Feature Extraction} \label{tumorextraction}
We applied our mitosis detection and tumor localization methods to identify biologically salient features in WSIs on both a patch-based and a whole slide level. Each approach allowed for extraction of features with varying granularities. 

Specifically, we preferentially selected fifty patches from the fringes of localized tumor regions with the largest area. Each patch, a 1k$\times$1k tissue sample at magnification level $40\times$, was input to our mitosis detectors which produced heatmaps of corresponding size identifying mitotic figure probability in the input image. We additionally represented each WSI with a comprehensive heatmap depicting mitotic figures in \textit{all} tumor regions. Both individual patches and the WSI heatmap are used to compute biological and data-driven mitosis features. 



\xhdr{Biological Features} From each selected patch, we extracted fifty morphometric and intensity based features to characterize biologically salient structural mitosis components. These features describe compositional and formational patterns that pathologists might observe. In addition, we characterized the distribution of mitoses throughout the entire WSI with sixty architectural features. Particularly, we analyzed the sparsity of mitosis distribution and second-order attributes including kurtosis, entropy, and skewness, providing a high-level interpretation of proliferative activity.

\xhdr{Data-Driven Features} Within each magnified patch, we additionally computed abstract deep learning-based features that represent attributes from learned filters. We segmented a $63\times63$ tissue patch around each identified mitosis for input to our mitosis detection networks. Each mitotic figure was subsequently characterized by 4,096 attributes to describe mitosis-specific structural minutiae. As each patch consisted of $N_i$ mitotic figures and was thereby associated with $4096 \times N_i$ distinct computational attributes, feature standardization was performed. We conducted post-processing $k$-means clustering on all individual mitosis feature vectors (of length 4,096) from every WSI patch in a 200-dimensional vector space. Each vector was associated with a cluster label $\in [1, 200]$ identifying its most similar sub-space. Finally, each WSI was distilled into a 200-bin histogram with frequencies corresponding to the cluster labels of each mitotic region within extracted patches, resulting in a fixed data-driven feature vector of length 200. 

All characteristic features extracted from tumor localization results (\ref{tumorlocalization}), general extraction methods (\ref{generalextraction}), and both deep learning and biological tumor-specific mitotic attributes (\ref{tumorextraction}) were used to predict the severity of tumor proliferation.



\section{Experiments and Results}
\subsection{Performance Evaluation}
\begin{figure}
\centering
\includegraphics[width=.5\textwidth]{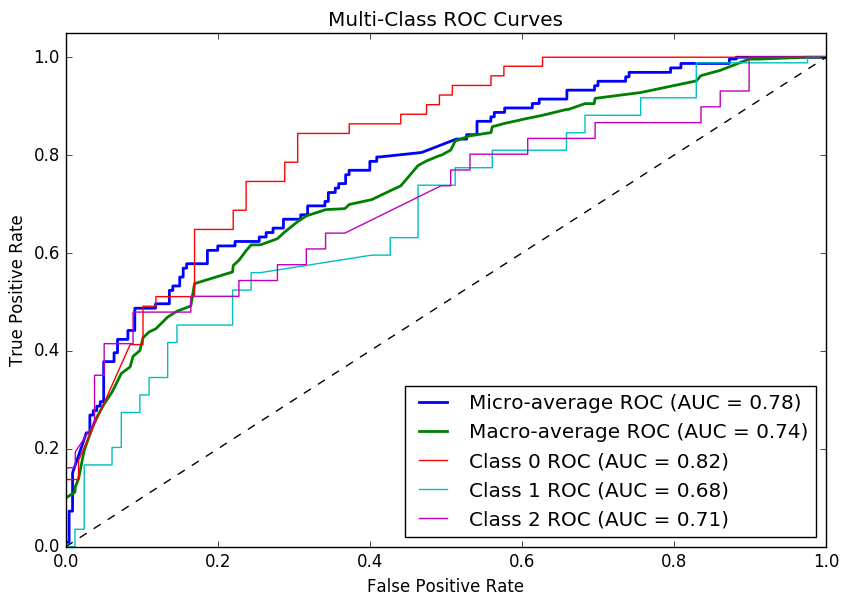}
\caption{\textit{Receiver Operating Characteristic Curves}. A depiction of the one \textit{vs} all predicted class based ROC curves with respective AUC values. }
\label{roc}
\end{figure}

\xhdr{Categorical Tumor Severity} A receiver operating characteristic (ROC) curve \cite{hanley1982meaning} detailing the ratio of true positives and false positives at varying thresholds is depicted in Fig. \ref{roc}. Each class was predicted in a one \textit{vs} all manner with mean values determined in five-fold cross-validation. The resulting micro-average AUROC of 0.78 validates our overall $f$-measure of 0.62, establishing the model's powerful discriminative potential among the three classes. Our method additionally achieved an accuracy of 0.72 (95\% CI: 0.67, 0.76) when compared to pathologist severity gradation, indicating marginal deviation of our predictions from the inter-pathologist agreement of 0.79 (95\% CI: 0.70, 0.85) \cite{veta2016mitosis}.

\begin{figure}
\centering
\includegraphics[width=0.46\textwidth]{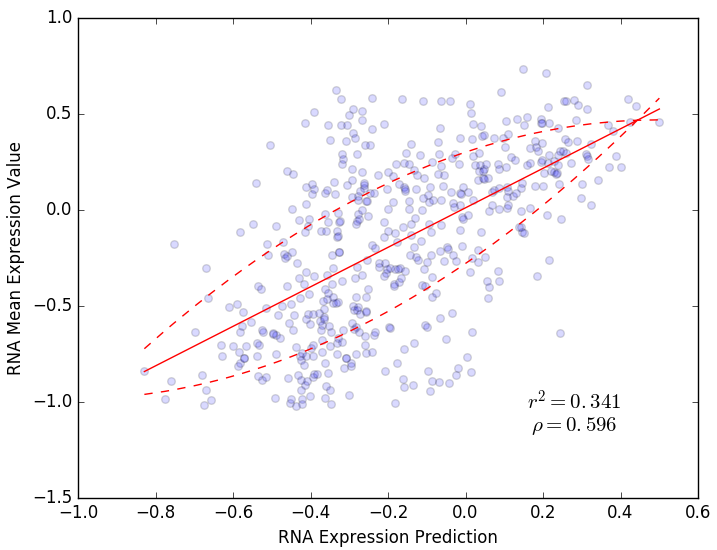}
\caption{\textit{Molecular Expression Prediction Results}. A depiction of the correlation between pipeline RNA expression prediction ($x$-axis) and calculated RNA expression from a full biopsy.}
\label{scatterplot}
\end{figure}

\xhdr{Molecular RNA Expression} Our best-performing regression model achieved a mean squared error of 0.119. Fig. \ref{scatterplot} depicts the correlation between our regression predictions and the calculated mean expression of eleven prognostic RNA strands. Our model, the first ever to predict gene expression data from histopathological image slides, achieves a Pearson's correlation coefficient value $r = 0.58$ and a Spearman's rank coefficient $\rho = 0.60$ ($p< 0.001$). 
Along with the low MSE, these results indicate the ability of our pipeline not only to match categorical pathologist diagnosis but also to provide significantly more salient information regarding the biological underpinnings defining the severity of tumor tissue. The overall performance metrics indicate the prognostic potential of our model in effectively evaluating histopathological slides without a molecular examination. 

\subsection{Specific Biomarker Analysis}
Our study further elucidated important biomarkers (regression features with $p < 0.005$) most important for proliferation evaluation. 
The number of mitoses, currently the sole feature used to manually diagnose tumor proliferation, was confirmed as relevant. Additionally, the standard deviation of the nucleus area of all identified mitoses was implicated. This attribute, known to characterize malignant neoplasms and dysplasia, was recently associated with breast cancer diagnosis \cite{muhammadnejad2013correlation}. 

In addition, the $p$-values and predictive significance of several new biomarkers suggest that they are significantly related to the progression of breast cancer. The mean mitotic eccentricity over each WSI, a feature characteristic of the development of a cleavage furrow in mitotic figures \cite{rappaport1982cytokinesis}, was found to be relevant, suggesting the differential importance of cytokinetic figures in diagnosis.  
The importance of compositional features of tumor tissue structures suggested prognostic information embedded in specific forms of tissue structure across the entire WSI. 
In addition to these interpretable biological features, low-level configurational and formational patterns identified by MitosNet were deemed relevant, denoting differential mitotic stages as prognostically significant. 

\section{Conclusion}
This study presented the first completely data-driven approach to develop and integrate numerous biologically salient classifiers into a single invasive breast cancer prognostic model. This model was used to predict tumor growth on a categorical and molecular scale and to discover novel image-related biomarkers critical to disease diagnosis. With our prediction framework performing equivalent to pathologist grading  and capturing the underlying complexity presented within tissue structure, early and less costly diagnoses for invasive breast cancer may allow for more effective and targeted treatments in clinical practice. 



\section*{Acknowledgments}
The authors would like to thank Babak Ehteshami Benjordi, Ben Glass, and Francisco Beca for their invaluable input. 
\bibliographystyle{IEEEtran}
\bibliography{bibliography}

\end{document}